\documentclass[10pt,twocolumn,letterpaper]{article}

\usepackage{cvpr}
\usepackage{times}
\usepackage{epsfig}
\usepackage{graphicx}
\usepackage{amsmath}
\usepackage{amssymb}

\usepackage{multirow}
\usepackage{array}
\usepackage{dsfont}
\usepackage{stfloats}
\usepackage{authblk}
\usepackage[pagebackref=true,breaklinks=true,letterpaper=true,colorlinks,bookmarks=false]{hyperref}
 
\cvprfinalcopy 

\def\cvprPaperID{1090}

\ifcvprfinal\fi

\begin{document}

\title{Learning Structured Inference Neural Networks with Label Relations}

\author[1,2]{Hexiang Hu}
\author[1]{Guang-Tong Zhou}
\author[1]{Zhiwei Deng}
\author[2]{Zicheng Liao}
\author[1]{Greg Mori}

\affil[1]{School of Computing Science, Simon Fraser University \\
Burnaby, BC, Canada}
\affil[2]{College of Computer Science and Technology, Zhejiang University \\
Hangzhou, Zhejiang, China}
\affil[ ]{\tt\small {\{hexiangh, gza11, zhiweid\}@sfu.ca}, {zliao@zju.edu.cn}, {mori@cs.sfu.ca} }
\maketitle

\begin{abstract}
Images of scenes have various objects as well as abundant attributes, and diverse levels of visual categorization are possible. A natural image could be assigned with fine-grained labels that describe major components, coarse-grained labels that depict high level abstraction, or a set of labels that reveal attributes. Such categorization at different concept layers can be modeled with label graphs encoding label information. In this paper, we exploit this rich information with a state-of-art deep learning framework, and propose a generic structured model that leverages diverse label relations to improve image classification performance. Our approach employs a novel stacked label prediction neural network, capturing both inter-level and intra-level label semantics. We evaluate our method on benchmark image datasets, and empirical results illustrate the efficacy of our model.
\end{abstract}

\section{Introduction}
\label{sec:intro}

\begin{figure}[htb]
\begin{center}
   \includegraphics[width=0.475\textwidth]{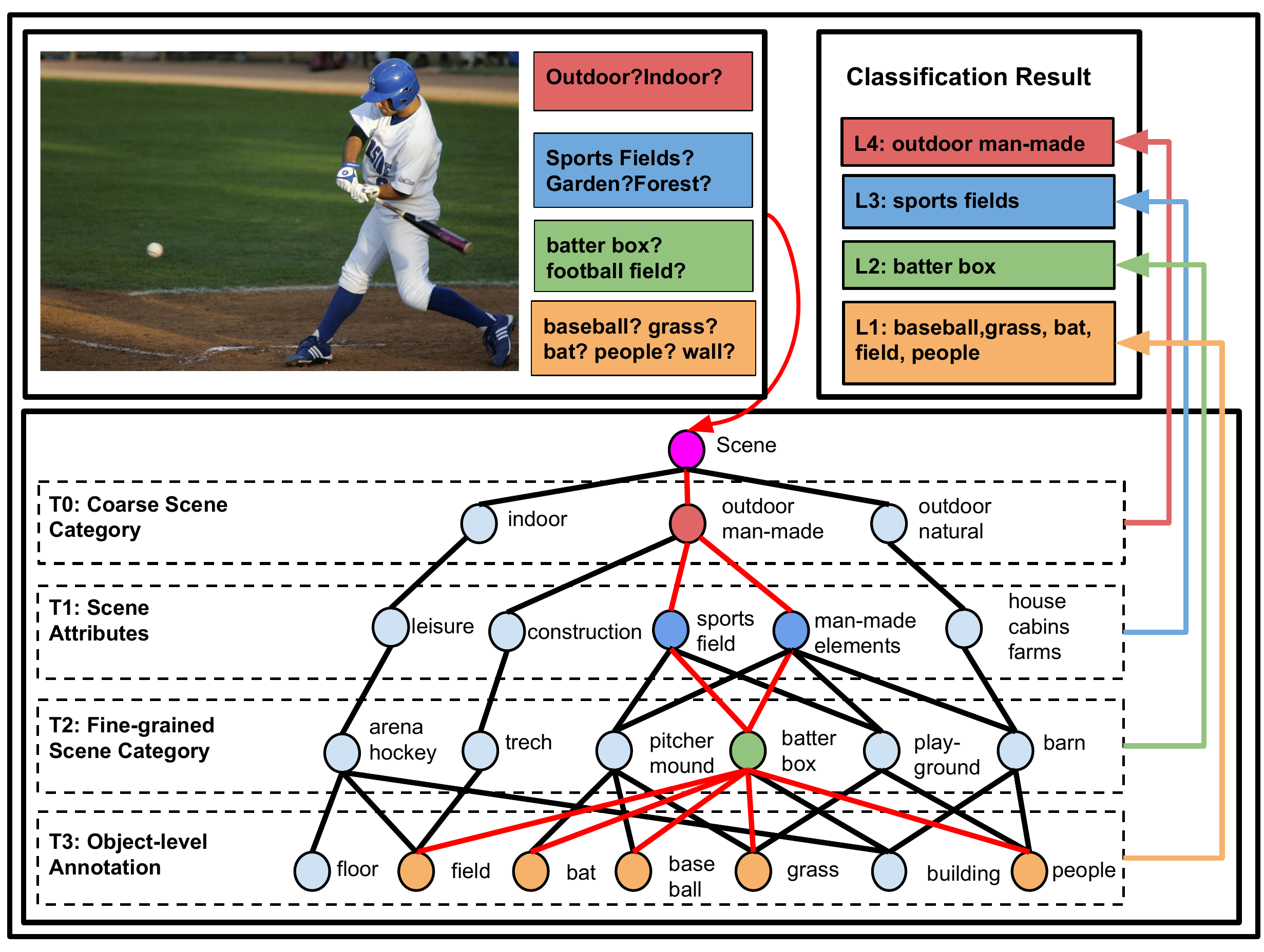}
\end{center}
\caption{
This image example has visual concepts at various levels, from sports field at high level to baseball and person at lower level. Our model leverages label relations and jointly predicts layered visual labels from an image using a structured inference neural network. In the graph, colored nodes correspond to the labels associated with the image, and red edges encode label relations.
}
\label{fig:intro}
\end{figure}

Standard image classification is a fundamental problem in computer vision -- assigning category labels to images. It can serve as a building block for many different computer vision tasks including object detection, visual segmentation, and scene parsing. Recent progress in deep learning~\cite{krizhevsky2012imagenet,sermanet2013overfeat,simonyan2014very,Szegedy15} significantly improved classification performance on large scale image datasets~\cite{ILSVRC15,xiao2010sun,nus-wide-civr09,lampert2014attribute}. Approaches typically assume image labels to be semantically independent and adapt either a multi-class or binary classifier to label images. In recent work~\cite{deng2014large,ding2015probabilistic}, deep learning methods that take advantage of label relations have been proposed to improve image classification performance.

However, in realistic settings, these label relationships could form a complicated graph structure. Take Figure~\ref{fig:intro} as an example. Various levels of interpretation could be formed to represent such an image.  This image of a \textit{baseball} scene could be described as an \emph{outdoor} image at coarse level, or with a more concrete concept such as \emph{sports field}, or with even more fine-grained labels such as \emph{batter's box} and objects such as \emph{grass}, \emph{bat}, \emph{person}.

Models that incorporate semantic label relationships could be utilized to generate better classification results. The desiderata for these models include the ability to model label-label relations such as positive or negative correlation, respect multiple concept layers obtainable from sources such as WordNet, and to handle partially observed label data -- given a subset of accurate labels for this image, infer the remaining missing labels.

The contribution of this paper is in developing a structured inference neural network that permits modeling complex relations between labels, ranging from hierarchical to within-layer dependencies.
We do this by defining a network in which a node is activated if its corresponding label is present in an image. We introduce stacked layers among these label nodes. These encode layer-wise connectivity among label classification scores, representing dependency from top-level coarse labels to bottom-level fine-grained labels. Activations are propagated bidirectionally and asynchronously on the label relation graph, passing information about the labels within or across concept layers to refine the labeling for the entire image.

We have evaluated our method on three image classification datasets (AWA dataset~\cite{lampert2014attribute},  NUS-WIDE dataset~\cite{nus-wide-civr09} and SUN397 dataset~\cite{xiao2010sun}). Experimental results show a consistent and significant performance gain with our structured label relation model compared with baseline and related methods.

\begin{figure*}[htb]
\begin{center}
    \includegraphics[width=1.0\textwidth] {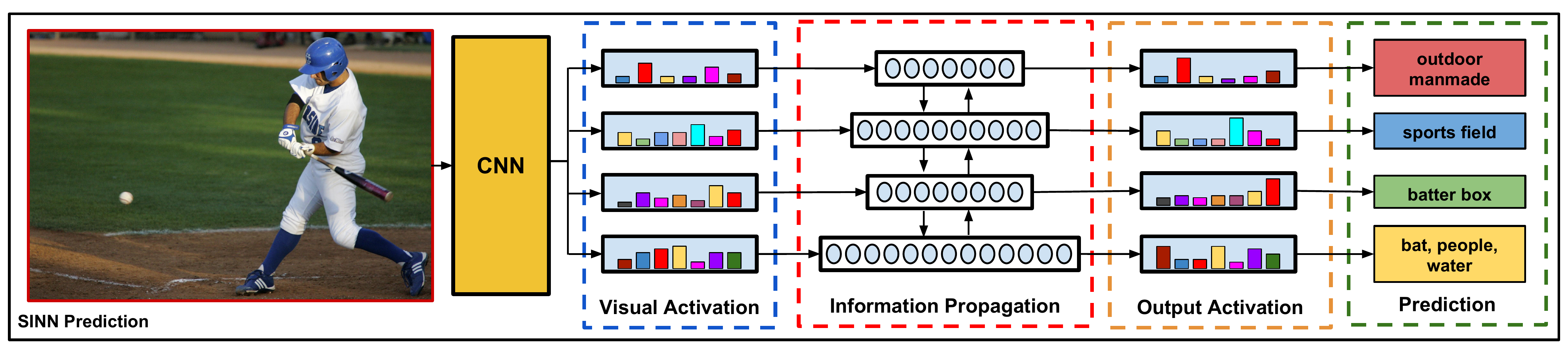}
\end{center}
\caption{The label prediction pipeline. Given an input image, we extract CNN features at the last fully connected layer as activation (in blue box) at different visual concept layers. We then propagate the activation information in a label (concept) relation graph through our structured inference neural network (in red box). The final label prediction (in green box) is made from the output activations (in yellow box) obtained after the inference process.}
\label{fig:pipeline}
\end{figure*}

\section{Related Work}
\label{sec:relatedwork}

Multi-level labeling of images has been addressed in a number of frameworks. In this section we review relevant work within probabilistic, max-margin, multi-task, and deep learning.

\noindent{\bf Structured label prediction with external knowledge}: Structured prediction approaches exist~\cite{Taskar03max,Tsochantaridis:2005:LMM}, in which a set of class labels are predicted jointly under a fixed loss function. Traditional approaches learn graph structure as well as associated weights that best explain the training data (\eg, \cite{deng2011fast}). When external knowledge of label relations (\eg, a taxonomy) is available, it is beneficial to integrate this knowledge to guide the traditional supervised learning systems. For example, Grauman \etal~\cite{grauman2011learning} and Hwang \etal~\cite{hwang2012semantic} took the WordNet category taxonomy to improve visual recognition. Johnson \etal~\cite{johnson2015love} and McAuley and Leskovec~\cite{mcauley2012image} used metadata from a social network to improve image classification. Ordonez \etal~\cite{ordonez2013large} leveraged associated image captions (words of ``naturalness'') to estimate entry-level labels of visual objects.

\noindent{\bf Multi-label classification with label relations}: Traditional multi-label classification cannot avoid predicting an image as both \textit{cat} and \textit{dog}, or an image as \textit{carnation} but not \textit{flower}.  Using external knowledge of label relations, Deng \etal~\cite{deng2014large} proposed a representation, the HEX graph, to express and enforce exclusion, inclusion and overlap relations between labels in multi-label classification. This model was further extended for ``soft'' label relations using the Ising model by Ding \etal~\cite{ding2015probabilistic}.

\noindent{\bf Structured model with convolutional neural networks (CNNs)}:
Structured deep models extend traditional CNNs to applications of structured label prediction, for which the CNN model is found insufficient to learn implicit constraints or structures between labels. Structured deep learning therefore jointly learns a structured model with the CNN framework. For example, for human pose estimation, Tompson \etal~\cite{tompson2014joint} take the CNN predictions as unary potentials for body parts and feed them to a MRF-like spatial model, which further learns pairwise potentials of part relations. Schwing and Urtasun~\cite{schwing2015fully} proposed a structured deep network by concatenating a densely connected MRF model to a CNN for semantic image segmentation, in which the CNN provides unary potentials as the MRF model imposes smoothness. Deng et. al.~\cite{deng2015structure} proposed a recurrent network that jointly performs message passing-style inference and learns graph structure for group activity recognition.

Our work combines these lines of work. We take the WordNet taxonomy as our external knowledge, expressing it as a label relation graph, and learning the structured labels within a deep network framework. Our contribution is in proposing a learning and inference algorithm that facilitates knowledge passing in the deep network based on label relations.

\noindent{\bf Multi-task joint learning}: Multi-task learning follows the same spirit of structured label prediction, with the distinction that the outputs of multiple (different but related) tasks are estimated.
Common jointly modeled tasks include segmentation and detection~\cite{Kumar2005,Wu2007}, segmentation and pose estimation~\cite{Kohli:2008}, or segmentation and object classification~\cite{Leibe04combinedobject}. An emerging topic of joint learning is in image understanding and text generation by leveraging intra-modal correspondences between visual and human language~\cite{KongCVPR14, karpathy2014deep}.

Our work can be naturally extended to multi-task learning, for which each layer of our model represents one task and the labels do not necessarily form a layered structure. Notably, we can improve existing multi-task learning methods by importing knowledge of intra-task label relations.

\section{Method}
\label{sec:method}

Our model jointly classifies images in a layered label space with external label relations.
The goal is to leverage the label relations to improve inference over the layered visual concepts.

We build our model on top of a state-of-the-art deep learning platform: given an image, we first extract CNN features from Krizhevsky \etal~\cite{krizhevsky2012imagenet} as visual activations at each concept layer. 
Concept layers are stacked from fine-grained to coarser levels. Label relations are defined between consecutive layers and form a layered graph. Inference over the label relation graph is inspired by the recent success of Recurrent Neural Networks (RNNs)~\cite{Hochreiter97,schuster1997bidirectional}, where we treat each concept layer as a timestep of a RNN. We connect neighboring timesteps to reflect the inter-layer label relations, while capturing intra-layer relations within each timestep. The label activations are propagated bidirectionally and asynchronously in the label relation graph to refine labeling for the given image. Figure~\ref{fig:pipeline}  shows an overview of our classification pipeline.

We denote the collection of training images as $\{I^{i}\}_{i=1}^{N}$, each with ground-truth label in every concept layer. We denote the labels of image $I^{i}$ as $\{y_{t}^{i}\}_{t=1}^{T}$, where $T$ is the total number of concept layers. And each concept layer $t$ has $n_{t}$ labels. The CNN framework of Krizhevsky \etal~\cite{krizhevsky2012imagenet} transforms each image $I^{i}$ into a 4096-dimensional feature vector, denoted as $CNN(I^{i})$.

\subsection{Learning Framework}
\label{sec:method:learn}

It is straightforward to build an image classification model, by adding a loss layer on top of the CNN features for each concept layer. Specifically, the activations on concept layer $t$ are computed as
\begin{equation}
\label{eqn:visual_activation}
    x_{t}^{i} = W_{t} \cdot CNN(I^{i}) + b_{x, t},
\end{equation}
where $W_{t} \in \mathbb{R}^{n_{t} \times 4096}$ and $b_{x, t} \in \mathbb{R}^{n_{t} \times 1}$ are linear transformation parameters and biases to classify the $n_{t}$ labels at concept layer $t$. Note that $x_{t}^{i} \in \mathcal{R}^{n_{t} \times 1}$ provides visual activation depending purely on the  image $I^{i}$. To generate label-specific probabilities, we can simply apply a sigmoid function (\ie, $\sigma{(z)} = \frac{1}{1 + e^{-z}}$) on the elements of $x_{t}^{i}$.

This classification model does not accommodate label relations within or across concept layers.
To leverage the benefit of label relations, we adopt an RNN-like inference framework. In the following, we first describe a top-down inference model, then a bidirectional inference model, and lastly propose our Structured Inference Neural Network, the SINN.

\subsubsection{Top-Down Inference Neural Network}
\label{sec:method:learn:rnn}

Our model is inspired by the recent success of RNNs~\cite{Graves08,Sak14}, which make use of dynamic sequential information in learning. RNNs are called \textit{recurrent} models because they perform the same computation for every timestep, with the input dependent on the current inputs and previous outputs. We apply a similar idea to our layered label prediction problem: we consider each concept layer as an individual timestep, and model the label relations within and across concept layers in the recurrent learning framework.

Specifically, at each timestep $t$, we compute an image $I^{i}$'s activations $a_{t}^{i} \in \mathbb{R}^{n_{t} \times 1}$ based on two terms: $a_{t-1}^{i} \in \mathbb{R}^{n_{t-1} \times 1}$, which are the activations from the last timestep $t-1$, and $x_{t}^{i} \in \mathbb{R}^{n_{t} \times 1}$, which are the activations from Eq.~\eqref{eqn:visual_activation}. The message passing process is defined as:
\begin{equation}
\label{eqn:rnn}
    a_{t}^{i} = V_{t-1,t} \cdot a_{t-1}^{i} + H_{t} \cdot x_{t}^{i} + b_{a, t},
\end{equation}
where $V_{t-1,t} \in \mathbb{R}^{n_{t} \times n_{t-1}}$ are the inter-layer model parameters capturing the label relations between two consecutive concept layers in top-down order, $H_{t} \in \mathbb{R}^{n_{t} \times n_{t}}$ are the intra-layer model parameters to account for the label relations within each concept layer, and $b_{a, t} \in \mathbb{R}^{n_{t} \times 1}$ are the model biases. A sigmoid function can be applied to $a_{t}^{i}$ to obtain label-specific prediction probabilities for image $I^{i}$.

Note that the inference process in Eq.~\eqref{eqn:rnn} is different from the standard RNN learning: Eq.~\eqref{eqn:rnn} unties $V_{t-1,t}$ and $H_{t}$ in each timestep, while the standard RNNs learn the same $V$ and $H$ parameters over and over on all timesteps.

To learn the model parameters $V$'s and $H$'s, we apply a sigmoid function function $\sigma$ on the activations $a_{t}^{i}$, and minimize the logistic cross-entropy loss with respect to $V$'s and $H$'s:
\begin{eqnarray}
\label{eqn:loss}
    E(\{a_{t}^{i}\}) & = & \sum_{i=1}^{N} \sum_{t=1}^{T} \sum_{y=1}^{n_{t}} \Big( \mathds{1}(y_{t}^{i} = y) \cdot \log\big(\sigma{(a_{t}^{i})}\big) \nonumber\\
    & + & \mathds{1}(y_{t}^{i} \neq y) \cdot \log\big(1 - \sigma{(a_{t}^{i})}\big) \Big),
\end{eqnarray}
where $\mathds{1}(z)$ is an indicator function which returns 1 if $z$ is true and 0 otherwise.

\subsubsection{BINN: Bidirectional Inference Neural Network}
\label{sec:method:learn:binn}

It makes more sense to model bidirectional inferences, as a concept layer is related to the two connected layers equally well. Therefore, we adopt the idea of bidirectional recurrent neural network~\cite{schuster1997bidirectional}, and propose the following bidirectional inference model:
\begin{eqnarray}
    \overrightarrow{a}_{t}^{i} & = & \overrightarrow{V}_{t-1,t} \cdot \overrightarrow{a}_{t-1}^{i} + \overrightarrow{H}_{t} \cdot x_{t}^{i} + \overrightarrow{b}_{t}, \label{eqn:binn:right}\\
    \overleftarrow{a}_{t}^{i} & = & \overleftarrow{V}_{t+1,t} \cdot \overleftarrow{a}_{t+1}^{i} + \overleftarrow{H}_{t} \cdot x_{t}^{i} + \overleftarrow{b}_{t},\label{eqn:binn:left}\\
    a_{t}^{i} & = & \overrightarrow{U}_{t} \cdot \overrightarrow{a}_{t}^{i} + \overleftarrow{U}_{t} \cdot \overleftarrow{a}_{t}^{i} + b_{a, t}.\label{eqn:binn:sum}
\end{eqnarray}

where Eqs.~\eqref{eqn:binn:right} and \eqref{eqn:binn:left} proceed as top-down propagation and bottom-up propagation, respectively, 
and Eq.~\eqref{eqn:binn:sum} aggregates the top-down and bottom-up messages into final activations for label prediction.
Here $\overrightarrow{U}_{t} \in \mathbb{R}^{n_{t} \times n_{t}}$ and $\overleftarrow{U}_{t} \in \mathbb{R}^{n_{t} \times n_{t}}$ are aggregation model parameters, and we use the arrows $\rightarrow$ and $\leftarrow$ to indicate the directions of label propagation

As in the top-down inference model, the bidirectional inference model captures both inter-layer and intra-layer label relations in the model parameters $V$'s and $H$'s. For inter-layer relations, we connect a label in one concept layer to any label in its neighboring concept layers. For intra-layer relations, we model fully-connected relations within each concept layer. The model parameters $V$'s, $H$'s and $U$'s are learned by minimizing the cross-entropy loss defined in Eq.~\eqref{eqn:loss}.

\subsubsection{SINN: Structured Inference Neural Network}
\label{sec:method:learn:sinn}

The fully connected bidirectional model is capable of representing all types of label relations.  In practice, however, it is hard to train a model on limited data due to the large number of free parameters.  To avoid this problem, we use a structured label relation graph to restrict information propagation.

We use structured label relations of positive correlation and negative correlation as prior knowledge to refine the model. Here is the intuition: since we know that \textit{office} is an \textit{indoor} scene, \textit{beach} is an \textit{outdoor} scene, and \textit{indoor} and \textit{outdoor} are mutually exclusive, a high score on \textit{indoor} should increase the probability of label \textit{office} and decrease the probability of label \textit{beach}. Labels that are not semantically related, e.g. motorcycle and shoebox, should not affect each other. The structured label relations can be obtained from semantic taxonomies, or by parsing WordNet relations~\cite{miller1995wordnet}. We describe the details of extracting label relations in Section~\ref{sec:experiment}.

\begin{figure}[htb]
\begin{center}
\includegraphics[width=0.475\textwidth]{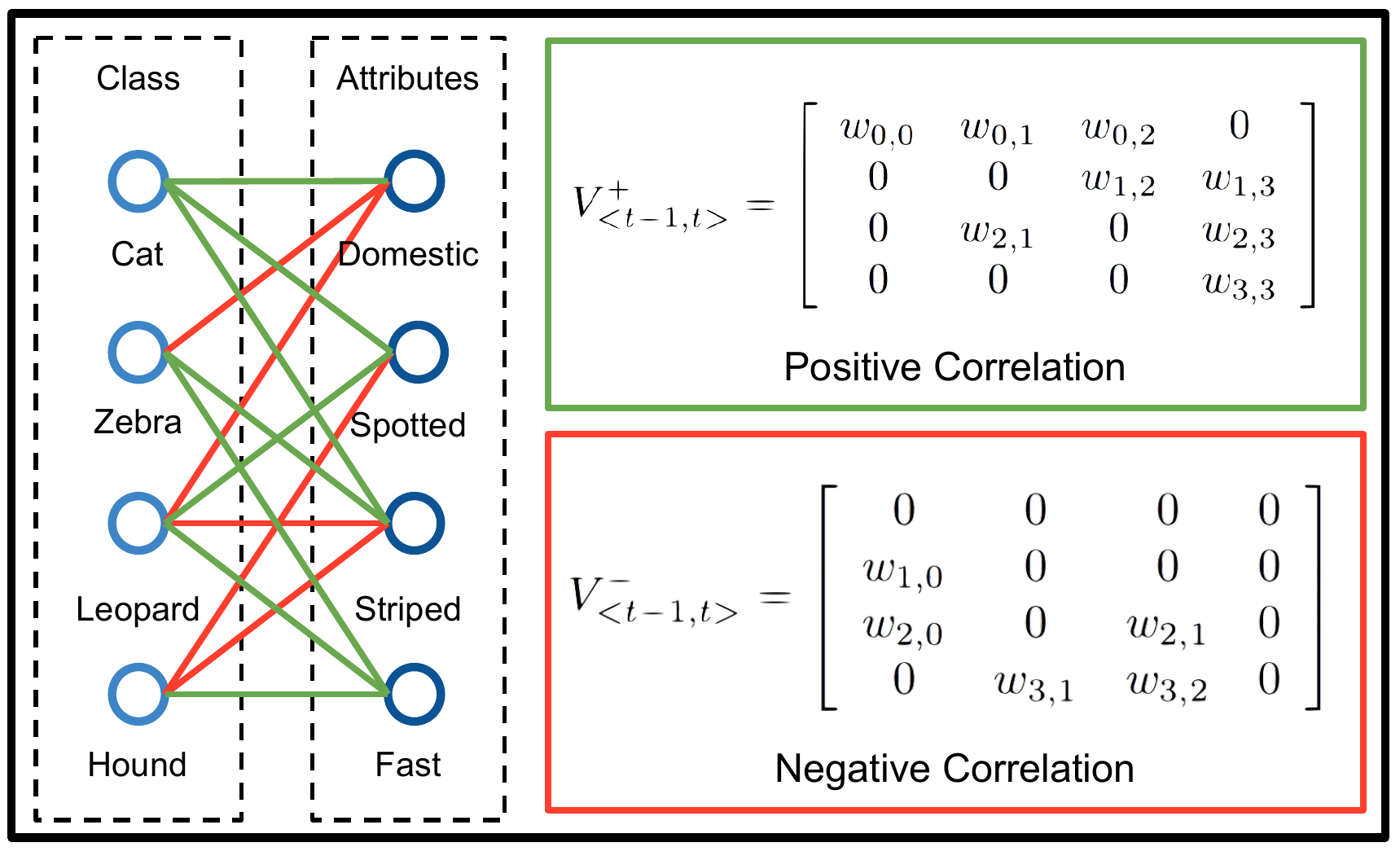}
\end{center}
\caption{An example showing the model parameters $V^{+}$ and $V^{-}$ between the \textit{animal} layer and the \textit{attribute} layer. Green edges in the graph represent positive correlation, and red edges represent negative correlation.}
\label{fig:structured_relation}
\end{figure}

We introduce the notation $V^{+}$, $V^{-}$, $H^{+}$ and $H^{-}$ to explicitly capture structured label relations in between and within concept layers, where the superscripts $+$ and $-$ indicate positive and negative correlation, respectively. These model parameters are masked metrics capturing the label relations. Instead of learning full parametrized metrics of $V^{+}$, $V^{-}$, $H^{+}$ and $H^{-}$, we freeze some elements to be zero if there is no semantic relation between the corresponding labels. For example, $V^{+}$ models the positive correlation in between two concept layers: only the label pairs that have positive correlation have learnable model parameters, while the rest are zeroed out to remove potential noise. A similar setting goes to $V^{-}$, $H^{+}$ and $H^{-}$. Figure~\ref{fig:structured_relation} shows an example positive correlation graph and a negative graph between two layers.

To implement the positive and negative label correlation, we propose the following structured message passing process:
\begin{eqnarray}
\label{eqn:sinn}
    \overrightarrow{a}_{t}^{i} & = & \gamma{(\overrightarrow{V}_{t-1,t}^{+} \cdot \overrightarrow{a}_{t-1}^{i})} + \gamma{(\overrightarrow{H}_{t}^{+} \cdot x_{t}^{i})} \label{eqn:sinn:right}\\
    & & - \gamma{(\overrightarrow{V}_{t-1,t}^{-} \cdot \overrightarrow{a}_{t-1}^{i})} - \gamma{(\overrightarrow{H}_{t}^{-} \cdot x_{t}^{i})} + \overrightarrow{b}_{t}, \nonumber\\
    \overleftarrow{a}_{t}^{i} & = & \gamma{(\overleftarrow{V}_{t+1,t}^{+} \cdot \overleftarrow{a}_{t+1}^{i})} + \gamma{(\overleftarrow{H}_{t}^{+} \cdot x_{t}^{i})} \label{eqn:sinn:left}\\
    & & - \gamma{(\overleftarrow{V}_{t+1,t}^{-} \cdot \overleftarrow{a}_{t+1}^{i})} - \gamma{(\overleftarrow{H}_{t}^{-} \cdot x_{t}^{i})} + \overleftarrow{b}_{t}, \nonumber\\
    a_{t}^{i} & = & \overrightarrow{U}_{t} \cdot \overrightarrow{a}_{t}^{i} + \overleftarrow{U}_{t} \cdot \overleftarrow{a}_{t}^{i} + b_{a, t}. \label{eqn:sinn:sum}
\end{eqnarray}

Here $\gamma{(\cdot)}$ stands for a ReLU activation function. It is essential for SINN as it enforces that activations from positive correlation always make positive contribution to output activation and keeps activations from negative correlation as negative contribution (notice the minus signs in Eqs~\eqref{eqn:sinn:right} and \eqref{eqn:sinn:left}). To learn the model parameters $V$'s, $H$'s, and $U$'s, we optimize the cross-entropy loss in Eq.~\eqref{eqn:loss}.

\subsection{Label Prediction}
\label{sec:method:prediction}

\begin{figure*}[htb]
\begin{center}
    \includegraphics[width=1.0\textwidth] {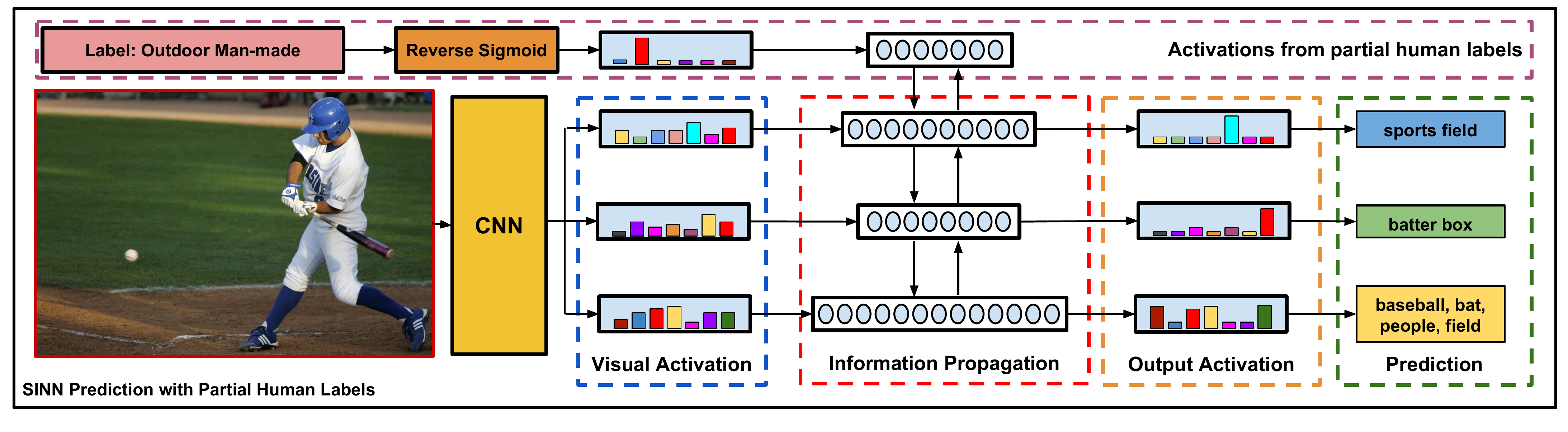}
\end{center}\vspace{-3mm}
\caption{The label prediction pipeline with partial observation. The pipeline is similar to Figure~\ref{fig:pipeline} except that we now have a partial observation that this image is \textit{outdoor man-made}. The SINN is able to take the observed label into consideration and improve the label predictions in the other concept layers.}
\label{fig:pipeline:observed}\vspace{3mm}
\end{figure*}

Now we introduce the method of predicting labels in test images with our model. As the model is trained with multiple concept layers, it is straightforward to recognize a label at each concept layer for the provided test image. This mechanism is called label prediction \textit{without observation} (the default pipeline shown in Figure~\ref{fig:pipeline}).

A more interesting application is to make predictions with \textit{partial observations} -- we want to predict labels in one concept layer given labels in another concept layers. Figure~\ref{fig:pipeline:observed} illustrates the idea. Given an image shown in the left side of Figure~\ref{fig:pipeline:observed}, we have more confidence to predict it as \textit{batter box} once we know it is an \textit{outdoor} image with attribute \textit{sports field}.

To make use of the partially observed labels in our SINN framework, we need to transform the observed binary labels into soft activation scores for SINN to improve the label prediction on the target concept layers. Recall that SINN minimizes cross-entropy loss which applies sigmoid functions on activations to generate label confidences. Thus, we reverse this process by applying the inverse sigmoid function on the binary ground-truth labels to obtain activations. Formally, we define the activation $a$ obtained from a ground-truth label $y$ as:
\begin{eqnarray}
\label{eqn:reverse_sigmoid}
g(y) & = &
\begin{cases}
    \log{\frac{y}{1 - (y + \epsilon)}},  \text{if } y = 0, \\
    \log{\frac{y}{1 - (y - \epsilon)}},  \text{if } y = 1.
\end{cases}
\end{eqnarray}
Note that we put a small perturbation $\epsilon$ on the ground-truth label $y$ for numerical stability. In our experiments, we set $\epsilon=0.001$.

\subsection{Implementation Details}
\label{sec:method:implement}
To optimize our learning objective, we use stochastic gradient descent with mini-batch size of 50 images and momentum of 0.9. For all training runs, we apply a two-stage policy as follows. In the first stage, we fixed pre-trained CNN networks, and train our SINN with a learning rate of 0.01 with fixed-size decay step. In the second stage, we set the learning rate as 0.0001 and fine-tune the CNN together with our SINN. We set the gradient clipping threshold to be 25 to prevent gradient explosion. The weight decay value for our training procedure is set to 0.0005.

In the computation of visual activations from the CNN, as different experiment datasets describe different semantic domains, we adopt different pretrained CNN models: ImageNet pretrained model~\cite{jia2014caffe} for experiments~\ref{sec:experiment:awa} and~\ref{sec:experiment:nuswide}, placenet pretrained model~\cite{zhou2014learning} for experiment~\ref{sec:experiment:sun397}.

\section{ Experiments }
\label{sec:experiment}

\begin{table*}[bp]
\small
\centering
\begin{tabular}{ |c|c|ccc| } \hline
Concept Layer & Method & {$MC Acc$} & $IoU Acc$ & $mAP_{L}$ \\
\hline\hline
\multirow{3}{*}{28 taxonomy terms}    & CNN + Logistics    & -                         & $80.41\pm{0.09}$       & $90.16\pm{0.10}$          \\
                                      & CNN + BINN         & -                         & $79.85\pm{0.13}$       & $89.92\pm{0.07}$          \\
                                      & CNN + SINN         & -                         & $\pmb{84.47\pm{0.38}}$ & $\pmb{93.00\pm{0.29}}$          \\
\hline\hline
\multirow{4}{*}{50 animal classes} & USE~\cite{NIPS2014_5289} + DECAF~\cite{donahue2013decaf}       & $46.42\pm{1.33}$ & - & -              \\
                                    & CNN + Logistics    & $78.44\pm{0.27}$          & $62.75\pm{0.26}$       & $78.35\pm{0.19}$            \\
                                    & CNN + BINN         & $79.00\pm{0.43}$          & $62.80\pm{0.25}$       & $78.88\pm{0.35}$            \\
                                    & CNN + SINN         & $\pmb{79.36\pm{0.43}}$    & $\pmb{66.60\pm{0.43}}$ & $\pmb{81.19\pm{0.14}}$             \\ 
\hline\hline
\multirow{3}{*}{85 attributes}      & CNN + Logistics    & -                         & $81.29\pm{0.10}$       & $93.29\pm{0.12}$            \\
                                    & CNN + BINN         & -                         & $80.64\pm{0.13}$       & $93.04\pm{0.13}$            \\
                                    & CNN + SINN         & -                         & $\pmb{86.92\pm{0.18}}$ & $\pmb{96.05\pm{0.07}}$            \\
\hline
\end{tabular}
\caption{Layered label prediction results on the AwA dataset.}
\label{tab:result:awa}
\end{table*}

We tested our method on three large-scale benchmark image datasets: the Animals with Attributes dataset (AwA)~\cite{lampert2014attribute}, the ADE20k dataset~\cite{nus-wide-civr09}, and the SUN397 dataset~\cite{xiao2010sun}. Each dataset has different concept layers and label relation graphs. Experimental results show that (1) our method effectively boosts classification performance using the label relation graphs; (2) our SINN model consistently outperforms baseline classifiers and related methods in all experiments; and (3) particularly, the SINN model achieves significant performance gain with partial human labels.

\noindent{\bf Dataset and Label relation generation} The AwA dataset contains an 85-attribute layer, a 50-animal-category layer and a 28-taxonomy-term layer. We extract the label relations from the WordNet taxonomy knowledge graph~\cite{grauman2011learning,hwang2012semantic,NIPS2014_5289}. 
The NUS-WIDE dataset is composed of Flickr images with 81 object category labels,  698 image group labels from image metadata, and 1000 noisy tags collected from users. We parse WordNet to obtain label similarity, and threshold the soft similarity values into positive and negative correlation for the label graph. 
The SUN397 dataset has a typical hierarchical structure in label space, with 397 fine-grained scene categories on the bottom layer, 16 general scene categories on middle layer, and 3 coarsest categories on the top. Here the label relations are also extracted from WordNet.

\noindent{\bf Baseline.} For each experiment, we compare our full method (CNN + SINN) with the baseline method: CNN + logistic regression. With further specifications, we may have extra baseline methods, such as CNN + BINN, CNN + logistic regression + extra tags, etc. We also compare our method with related state-of-the-art methods.

\noindent{\bf Evaluation metrics.} We measure classification performance by mean average precision ($mAP$) in all comparisons. $mAP$ is a widely used metric for label-based retrieval and ranking. It measures the averaged performance over all label categories. In addition to $mAP$, we also adopted various metrics for special cases.

In the case of NUS-WIDE, the task is multi-label classification. We adopt the setting of \cite{johnson2015love} and report mAP per label ($mAP_{L}$) and mAP per image ($mAP_{I}$) for easy comparison. For comparison with related works (~\cite{mcauley2012image,gong2013deep,johnson2015love}) on NUS-WIDE, we also compute the per image and per label precisions and recalls. We abbreviate these metrics as $Prec_{L}$ for precision per label, $Prec_{I}$ for precision per image, $Rec_{L}$ for recall per label, and $Rec_{I}$ for precision per image.

For AwA and SUN397, we also compute the multi-class accuracy ($MC Acc$) and the intersection-over-union accuracy ($IoU Acc$). $MC Acc$ is a standard measurement for image classification problems. It averages per class accuracies as the final result. $IoU Acc$ is a common prediction measurement for multi-label classification, based on the hamming distance of predicted labels to ground-truth labels.

\subsection{AwA: Layered Prediction with Label Relations}
\label{sec:experiment:awa}

This experiment demonstrates the label prediction capability of our SINN model and the effectiveness of adding structured label relations for label prediction. We run each method five times with five random splits -- 60\% for training and 40\% for test. We report the average performance as well as the standard deviation of each performance measure.

Note that there is very little related work with layered label prediction on AwA. The most relevant one is work by Hwang and Sigal~\cite{NIPS2014_5289} on unified semantic embedding (USE). The comparison is not strictly fair, as the train/test splits are different. Further, we include our BINN model without specifying the label relation graphs (see Section~\ref{sec:method:learn:binn}) as a baseline method in this experiment, as it can verify the performance gain in our model from including structure. The results are in Table~\ref{tab:result:awa}.

\noindent{\bf Results.} Table~\ref{tab:result:awa} shows that our method outperforms the baseline methods (CNN + Logistics and CNN + BINN variants) as well as the USE method, in terms of each concept layer and each performance metric. It validates the efficacy of our proposed model for image classification. Note that for the results in Table~\ref{tab:result:awa}, we did not fine-tune the first seven layers of CNN~\cite{krizhevsky2012imagenet} for fairer comparison with Hwang and Sigal~\cite{NIPS2014_5289} (which only makes use of DECAF features~\cite{donahue2013decaf}). Fine-tuning the first seven CNN layers further improves $IoU Acc$ at each concept layer to ${86.06\pm{0.72}}$ (28 taxonomy terms), ${69.17\pm{1.00}}$ (50 animal classes), ${88.22\pm{0.38}}$ (85 attributes), and $mAP_{L}$ to ${94.17\pm{0.55}}$ (28 taxonomy terms), ${83.12\pm{0.69}}$ (50 animal classes), ${96.72\pm{0.20}}$ (85 attributes), respectively.

\subsection{NUS-WIDE: Multi-label Classification with Partial Human Labels of Tags and Groups}
\label{sec:experiment:nuswide}

\begin{figure*}[tp]
\begin{center}
   \includegraphics[width=1.0\textwidth]{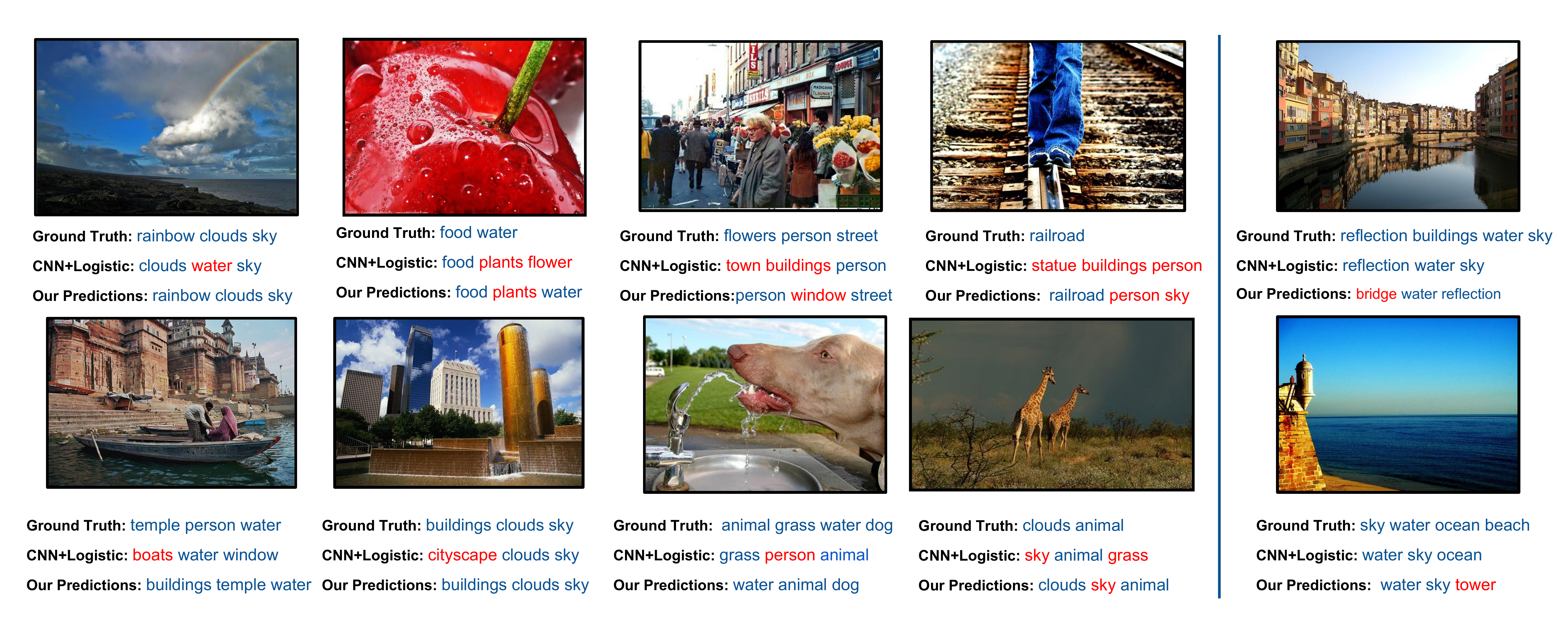}
\end{center}\vspace{-5mm}
\caption{Visualization results (best viewed in color). We pick up 10 representative images from NUS-WIDE, and visualize the predicted labels of our method compared with CNN + Logistics. Under each image, we provide the ground-truth labels for ease of reference, and list the top-3 highest scoring predicted labels for each compared method. Correct predictions are marked in blue and incorrect predictions are in red. Failure cases are shown in the rightmost column.}
\label{fig:viz}
\end{figure*}

\begin{table*}[bp]
\centering
\tabcolsep 5pt
\small
\begin{tabular}{|c|cc|cccc|}
\hline
Method & $mAP_L$ & $mAP_I$ & $Rec_L$ & $Prec_L$ & $Rec_I$ & $Prec_I$ \\
\hline\hline
{Graphical Model~\cite{mcauley2012image} }                      & $49.00$                & -                 & -                 & -                 & -                 & -                             \\
{CNN + WARP~\cite{gong2013deep} }                               & -                      & -                 & $35.60 $          & $31.65$           & $60.49$           & $48.59$                       \\ 
{5k tags + Logistics~\cite{johnson2015love}}                    & $43.88\pm{0.32}$       & $77.06\pm{0.14}$  & $47.52\pm{2.59}$  & $46.83\pm{0.89}$  & $71.34\pm{0.16}$  & $51.18\pm{0.16}$              \\
{Tag neighbors + 5k tags~\cite{johnson2015love}}                & $61.88\pm{0.36}$       & $80.27\pm{0.08}$  & $57.30\pm{0.44}$  & $54.74\pm{0.63}$  & $75.10\pm{0.20}$  & $53.46\pm{0.09}$              \\
\hline\hline
{CNN  + Logistics}                                              & $46.94\pm{0.47}$       & $72.25\pm{0.19}$  & $45.03\pm{0.44}$  & $45.60\pm{0.35}$  & $70.77\pm{0.21}$  & $51.32\pm{0.14}$              \\ 
{1k tags + Logistics}                                           & $50.33\pm{0.37}$       & $66.57\pm{0.12}$  & $23.97\pm{0.23}$  & $47.40\pm{0.07}$  & $64.95\pm{0.18}$  & $47.40\pm{0.07}$              \\
{1k tags + Groups + Logistics}                                  & $52.81\pm{0.40}$       & $68.04\pm{0.12}$  & $25.54\pm{0.24}$  & $49.26\pm{0.15}$  & $65.99\pm{0.15}$  & $48.13\pm{0.05}$              \\
{1k tags + Groups + CNN + Logistics}                            & $54.67\pm{0.57}$       & $77.81\pm{0.22}$  & $50.83\pm{0.53}$  & $49.36\pm{0.30}$  & $75.38\pm{0.16}$  & $54.61\pm{0.09}$              \\
\hline\hline
{1k tags + CNN + SINN}                                          & $67.20\pm{0.60}$       & $81.99\pm{0.14}$  & $59.82\pm{0.12}$  & $57.02\pm{0.57}$  & $78.78\pm{0.13}$  & $56.84\pm{0.07}$              \\
{1k tags + Groups + CNN + SINN}                                 & $\pmb{69.24\pm{0.47}}$ & $\pmb{82.53\pm{0.15}}$   & $\pmb{60.63\pm{0.67}}$ & $\pmb{58.30\pm{0.33}}$ & $\pmb{79.12\pm{0.18}}$ & $\pmb{57.05\pm{0.09}}$ \\
\hline
\end{tabular}
\caption{Results on NUS-WIDE. We measure precision $Pre_L$, $Pre_I$ and recall $Rec_L$, $Rec_I$ with n = 3 labels for each image.}
\label{tab:result:nuswide}
\end{table*}

This experiment shows our model's capability to use noisy tags and structured tag-label relations to improve multi-label classification. The original NUS-WIDE dataset consists of 269,648 images collected from Flickr with 81 ground-truth concepts. As previous work used various evaluation metrics and experiment settings, and there are no fixed train/test splits, it is hard to make direct comparisons. Also note that a fraction of previously used images are unavailable now due to Flickr copyright.

In order to make our result as comparable as possible, we tried to set up the experiments according to previous work. We collected all available images and discard images with missing labels as previous work did~\cite{johnson2015love, gong2013deep}, and got 168,240 images of the original dataset. To make our result comparable with~\cite{johnson2015love}, we use 5 random splits with the same train/test ratio as~\cite{johnson2015love} -- there are 132,575 training images and 35,665 test images in each split.

To compare our method with~\cite{mcauley2012image, johnson2015love}, we also used the tags and metadata groups in our experiment. Different from their settings, instead of augmenting images with 5000 tags, we only used 1000 tags, and augment the image with 698 group labels obtained from image medatada to form a three-layer group-concept-tag graph. Instead of using the tags as sparse binary input features (as in~\cite{mcauley2012image,johnson2015love}), we convert them to observed labels and feed them to our model.

The baselines for comparison are as follows. As our usual baseline, we extract features from a CNN pretrained on ImageNet~\cite{ILSVRC15} and train a logistic classifier on top of it. In addition, we set up a group of baselines that make use of the groups and tags as binary indicator feature vectors for logistic regression. These baselines serve as the control group to evaluate the quality of metadata we used in SINN. Next, a stronger baseline that uses both CNN output and metadata vector with logistic classifier was evaluated. This method has a similar setting as that of the state-of-art method by Johnson \etal~\cite{johnson2015love}, with difference in visual feature (CNN on image in our method versus CNN on image neighborhood) and tag feature (1k tag vector versus 5k tag vector). 

We report our results on this dataset with two settings for our SINN, the first using 1k tags as the only observations to a bottom level of the relation graph. This method provides a good comparison to the tag neighborhood + tag vector \cite{johnson2015love}, as we did not use extra information other than tags. In the second setting, we make both group and tag levels observable to our SINN, which achieves the best performance. We also compared our results with that of McAuley \etal~\cite{mcauley2012image}, Gong \etal~\cite{gong2013deep}. The results are summarized in Table~\ref{tab:result:nuswide}. Note that we did not report our performance with fine-tuning the first seven layers of the CNN in this table, so as to make direct comparison of structured inference on SINN with our baseline method CNN + Logistics. Fine-tuned CNN with SINN improves $mAP_L$ to $70.01\pm{0.40}$ and $mAP_I$ to $83.68\pm{0.13}$.

\begin{table*}[tp]
\small
\tabcolsep 5pt
\centering
\begin{tabular}{ |c|c|ccc| } \hline
Concept Layer & Method & {$MC Acc$} & $IoU Acc$ & $mAP_{L}$ \\ \hline\hline
\multirow{3}{*}{3 coarse scene categories}        & {CNN + Logistics}       & -                             & $83.67\pm{0.18}$          & $95.19\pm{0.07}$          \\ 
                                            & {CNN + BINN}                  & -                             & $83.63\pm{0.24}$          & $95.19\pm{0.03}$          \\ 
                                            & {CNN + SINN}                  & -                             & $\pmb{85.95\pm{0.44}}$    & $\pmb{96.40\pm{0.18}}$    \\ \hline\hline
\multirow{3}{*}{16 general scene categories}& {CNN + Logistics}             & -                             & $64.30\pm{0.27}$          & $83.30\pm{0.19}$          \\ 
                                            & {CNN + BINN}                  & -                             & $63.40\pm{0.35}$          & $82.93\pm{0.14}$          \\ 
                                            & {CNN + SINN}                  & -                             & $\pmb{66.46\pm{1.10}}$    & $\pmb{84.97\pm{0.96}}$    \\ \hline\hline
\multirow{4}{*}{397 fine-grained scene categories}                 & {Image features + SVM~\cite{xiao2010sun, xiao2014sun}}      & $42.70$    & -            & -    \\
                                            & {CNN + Logistics}             & $\pmb{57.86\pm{0.38}}$        & $35.97\pm{0.37}$          & $55.31\pm{0.30}$          \\
                                            & {CNN + BINN}                  & $57.52\pm{0.29}$              & $35.44\pm{1.02}$          & $55.57\pm{0.63}$          \\
                                            & {CNN + SINN}                  & $57.60\pm{0.38}$              & $\pmb{37.71\pm{1.13}}$    & $\pmb{58.00\pm{0.33}}$    \\ \hline
\end{tabular}
\caption{Layered label prediction results on the SUN397 dataset.}
\label{tab:result:sun397}
\end{table*}

\begin{table}[bp]
\small
\tabcolsep 2pt
\centering
\begin{tabular}{ |c|cc| } \hline
Method                                                      & $MC Acc$                  & $mAP_{L}$                 \\ \hline\hline
{Image features + SVM~\cite{xiao2010sun, xiao2014sun}}      & $42.70$                   & -                         \\
{CNN + Logistics}                                           & $57.86\pm{0.38}$          & $55.31\pm{0.30}$          \\ 
{CNN + BINN}                                                & $57.52\pm{0.29}$          & $55.57\pm{0.63}$          \\
{CNN + SINN}                                                & $57.60\pm{0.38}$          & $58.00\pm{0.33}$          \\ \hline
\hline
{CNN + Logistics + Partial Labels}                           & $59.08\pm{0.27}$          & $56.88\pm{0.29}$          \\
{CNN + SINN + Partial Labels}                                & $\pmb{63.46\pm{0.18}}$    & $\pmb{64.63\pm{0.28}}$    \\ \hline
\end{tabular}
\caption{Recognition results on the 397 fine-grained scene categories. Note that the last two compared methods make use of partially observed labels from the top (coarsest) scene layer, \ie, \textit{indoor}, \textit{outdoor man-made}, and \textit{outdoor natural}.}
\label{tab:result:weak_label}
\end{table}

\noindent{\bf Results.} Table~\ref{tab:result:nuswide} shows that our proposed method outperforms all baseline methods and existing approaches (\eg, \cite{johnson2015love,gong2013deep,mcauley2012image}) by a large margin. Note that the results are not directly comparable due to different settings in train/test splits. However, the results show that, by modeling label relations between tags, groups and concepts, our model achieves dramatic improvement on visual prediction.

We visualize some results in Figure~\ref{fig:viz} showing exemplars on which our method improves over baseline predictions.

\subsection{SUN397: Improving Scene Recognition with and without partially Observed Labels} 
\label{sec:experiment:sun397}

We conducted two experiments on the SUN397 dataset. The first experiment is similar to the study on AwA: we applied our model to layered image classification with label relations, and compare our model with CNN + Logistics and CNN + BINN baselines, as well as a state-of-the-art approach~\cite{xiao2010sun,xiao2014sun}. For fair comparison, we used the same train/test split ratio as \cite{xiao2010sun, xiao2014sun}, where we have 50 training and test images in each of the 397 scene categories. To migrate the randomness in sampling, we also repeat the experiment 5 times and report the average performance as well as the standard deviations. The results are summarized in Table~\ref{tab:result:sun397}, showing that our proposed method again achieves a considerable performance gain over all the compared methods. 

In the second experiment, we considered partially observed labels from the top (coarsest) scene layer as input to our inference framework. In other words, we assume we know whether an image is \textit{indoor}, \textit{outdoor man-made}, or \textit{outdoor natural}. We compare the 397 fine-grained scene recognition performance in Table~\ref{tab:result:weak_label}. 
We compare to a set of baselines, including CNN + Logistics + Partial Labels, that considers the partial labels as an extra binary indicator feature vector for logistic regression. 
Results show that our method combined with partial labels (\ie, CNN + SINN + Partial Labels) improves over baselines, exceeding the second best by 4\% $MC Acc$ and 6\% $mAP_{L}$. 

\section{ Conclusion }
\label{sec:concl}

We have presented a structured inference neural network (SINN) for layered label prediction. Our model makes use of label relation graphs and concept layers to augment inference of semantic image labels. Beyond this, our model can be flexibly extended to consider partially observed human labels. We borrow the idea of RNNs to implement our SINN model, and combine it organically with an underlying CNN visual output. Experiments on three benchmark image datasets show the effectiveness of the proposed method in standard image classification tasks. Moreover, we also demonstrate empirically that label prediction is further improved once partially observed human labels are fed into the SINN.

\section*{Acknowledgements}

This work was supported by grants from NSERC and Nokia.

{\small
\bibliographystyle{ieee}
\bibliography{ref}

\begin{thebibliography}{10}\itemsep=-1pt

\bibitem{nus-wide-civr09}
T.-S. Chua, J.~Tang, R.~Hong, H.~Li, Z.~Luo, and Y.-T. Zheng.
\newblock Nus-wide: A real-world web image database from national university of
  singapore.
\newblock In {\em CIVR}, 2009.

\bibitem{deng2014large}
J.~Deng, N.~Ding, Y.~Jia, A.~Frome, K.~Murphy, S.~Bengio, Y.~Li, H.~Neven, and
  H.~Adam.
\newblock Large-scale object classification using label relation graphs.
\newblock In {\em ECCV}. 2014.

\bibitem{deng2011fast}
J.~Deng, S.~Satheesh, A.~C. Berg, and F.~Li.
\newblock Fast and balanced: Efficient label tree learning for large scale
  object recognition.
\newblock In {\em NIPS}, 2011.

\bibitem{deng2015structure}
Z.~Deng, A.~Vahdat, H.~Hu, and G.~Mori.
\newblock Structure inference machines: Recurrent neural networks for analyzing
  relations in group activity recognition.
\newblock 2016.

\bibitem{ding2015probabilistic}
N.~Ding, J.~Deng, K.~Murphy, and H.~Neven.
\newblock Probabilistic label relation graphs with ising models.
\newblock {\em ICCV}, 2015.

\bibitem{donahue2013decaf}
J.~Donahue, Y.~Jia, O.~Vinyals, J.~Hoffman, N.~Zhang, E.~Tzeng, and T.~Darrell.
\newblock Decaf: A deep convolutional activation feature for generic visual
  recognition.
\newblock {\em ICML}, 2013.

\bibitem{gong2013deep}
Y.~Gong, Y.~Jia, T.~Leung, A.~Toshev, and S.~Ioffe.
\newblock Deep convolutional ranking for multilabel image annotation.
\newblock {\em ICLR}, 2014.

\bibitem{grauman2011learning}
K.~Grauman, F.~Sha, and S.~J. Hwang.
\newblock Learning a tree of metrics with disjoint visual features.
\newblock In {\em NIPS}, 2011.

\bibitem{Graves08}
A.~Graves and J.~Schmidhuber.
\newblock Offline handwriting recognition with multidimensional recurrent
  neural networks.
\newblock In {\em NIPS}, 2008.

\bibitem{Hochreiter97}
S.~Hochreiter and J.~Schmidhuber.
\newblock Long short-term memory.
\newblock {\em Neural Computation (NC)}, 9(8):1735--1780, 1997.

\bibitem{hwang2012semantic}
S.~J. Hwang, K.~Grauman, and F.~Sha.
\newblock Semantic kernel forests from multiple taxonomies.
\newblock In {\em NIPS}, 2012.

\bibitem{NIPS2014_5289}
S.~J. Hwang and L.~Sigal.
\newblock A unified semantic embedding: Relating taxonomies and attributes.
\newblock In {\em NIPS}, 2014.

\bibitem{jia2014caffe}
Y.~Jia, E.~Shelhamer, J.~Donahue, S.~Karayev, J.~Long, R.~Girshick,
  S.~Guadarrama, and T.~Darrell.
\newblock Caffe: Convolutional architecture for fast feature embedding.
\newblock In {\em ACM MM}, 2014.

\bibitem{johnson2015love}
J.~Johnson, L.~Ballan, and F.-F. Li.
\newblock Love thy neighbors: Image annotation by exploiting image metadata.
\newblock {\em ICCV}, 2015.

\bibitem{karpathy2014deep}
A.~Karpathy and L.~Fei-Fei.
\newblock Deep visual-semantic alignments for generating image descriptions.
\newblock {\em CVPR}, 2015.

\bibitem{Kohli:2008}
P.~Kohli, J.~Rihan, M.~Bray, and P.~H. Torr.
\newblock Simultaneous segmentation and pose estimation of humans using dynamic
  graph cuts.
\newblock {\em International Journal of Computer Vision (IJCV)},
  79(3):285--298, 2008.

\bibitem{KongCVPR14}
C.~Kong, D.~Lin, M.~Bansal, R.~Urtasun, and S.~Fidler.
\newblock What are you talking about? text-to-image coreference.
\newblock In {\em CVPR}, 2014.

\bibitem{krizhevsky2012imagenet}
A.~Krizhevsky, I.~Sutskever, and G.~E. Hinton.
\newblock Imagenet classification with deep convolutional neural networks.
\newblock In {\em NIPS}, 2012.

\bibitem{Kumar2005}
M.~Kumar, P.~Ton, and A.~Zisserman.
\newblock Obj cut.
\newblock In {\em CVPR}, 2005.

\bibitem{lampert2014attribute}
C.~H. Lampert, H.~Nickisch, and S.~Harmeling.
\newblock Attribute-based classification for zero-shot visual object
  categorization.
\newblock {\em IEEE Transactions on Pattern Analysis and Machine Intelligence
  (TPAMI)}, 36(3):453--465, 2014.

\bibitem{Leibe04combinedobject}
B.~Leibe, A.~Leonardis, and B.~Schiele.
\newblock Combined object categorization and segmentation with an implicit
  shape model.
\newblock In {\em ECCV Workshop}, 2004.

\bibitem{mcauley2012image}
J.~McAuley and J.~Leskovec.
\newblock Image labeling on a network: using social-network metadata for image
  classification.
\newblock In {\em ECCV}. 2012.

\bibitem{miller1995wordnet}
G.~A. Miller.
\newblock Wordnet: a lexical database for english.
\newblock {\em Communications of the ACM (CACM)}, 38(11):39--41, 1995.

\bibitem{ordonez2013large}
V.~Ordonez, J.~Deng, Y.~Choi, A.~C. Berg, and T.~Berg.
\newblock From large scale image categorization to entry-level categories.
\newblock In {\em ICCV}, 2013.

\bibitem{ILSVRC15}
O.~Russakovsky, J.~Deng, H.~Su, J.~Krause, S.~Satheesh, S.~Ma, Z.~Huang,
  A.~Karpathy, A.~Khosla, M.~Bernstein, A.~C. Berg, and L.~Fei-Fei.
\newblock {ImageNet} large scale visual recognition challenge.
\newblock {\em International Journal of Computer Vision (IJCV)}, pages 1--42,
  2015.

\bibitem{Sak14}
H.~Sak, A.~W. Senior, and F.~Beaufays.
\newblock Long short-term memory recurrent neural network architectures for
  large scale acoustic modeling.
\newblock In {\em Interspeech}, 2014.

\bibitem{schuster1997bidirectional}
M.~Schuster and K.~K. Paliwal.
\newblock Bidirectional recurrent neural networks.
\newblock {\em IEEE Transactions on Signal Processing (TSP)},
  45(11):2673--2681, 1997.

\bibitem{schwing2015fully}
A.~G. Schwing and R.~Urtasun.
\newblock Fully connected deep structured networks.
\newblock {\em arXiv preprint arXiv:1503.02351}, 2015.

\bibitem{sermanet2013overfeat}
P.~Sermanet, D.~Eigen, X.~Zhang, M.~Mathieu, R.~Fergus, and Y.~LeCun.
\newblock Overfeat: Integrated recognition, localization and detection using
  convolutional networks.
\newblock {\em ICLR}, 2014.

\bibitem{simonyan2014very}
K.~Simonyan and A.~Zisserman.
\newblock Very deep convolutional networks for large-scale image recognition.
\newblock {\em arXiv preprint arXiv:1409.1556}, 2014.

\bibitem{Szegedy15}
C.~Szegedy, W.~Liu, Y.~Jia, P.~Sermanet, S.~Reed, D.~Anguelov, D.~Erhan,
  V.~Vanhoucke, and A.~Rabinovich.
\newblock Going deeper with convolutions.
\newblock In {\em CVPR}, 2015.

\bibitem{Taskar03max}
B.~Taskar, C.~Guestrin, and D.~Koller.
\newblock Max-margin markov networks.
\newblock In {\em NIPS}, 2003.

\bibitem{tompson2014joint}
J.~J. Tompson, A.~Jain, Y.~LeCun, and C.~Bregler.
\newblock Joint training of a convolutional network and a graphical model for
  human pose estimation.
\newblock In {\em NIPS}, 2014.

\bibitem{Tsochantaridis:2005:LMM}
I.~Tsochantaridis, T.~Joachims, T.~Hofmann, and Y.~Altun.
\newblock Large margin methods for structured and interdependent output
  variables.
\newblock {\em Journal of Machine Learning Research (JMLR)}, 6:1453--1484,
  2005.

\bibitem{Wu2007}
B.~Wu and R.~Nevatia.
\newblock Simultaneous object detection and segmentation by boosting local
  shape feature based classifier.
\newblock In {\em CVPR}, 2007.

\bibitem{xiao2014sun}
J.~Xiao, K.~A. Ehinger, J.~Hays, A.~Torralba, and A.~Oliva.
\newblock Sun database: Exploring a large collection of scene categories.
\newblock {\em International Journal of Computer Vision (IJCV)}, pages 1--20,
  2014.

\bibitem{xiao2010sun}
J.~Xiao, J.~Hays, K.~Ehinger, A.~Oliva, A.~Torralba, et~al.
\newblock Sun database: Large-scale scene recognition from abbey to zoo.
\newblock In {\em CVPR}, 2010.

\bibitem{zhou2014learning}
B.~Zhou, A.~Lapedriza, J.~Xiao, A.~Torralba, and A.~Oliva.
\newblock Learning deep features for scene recognition using places database.
\newblock In {\em NIPS}, 2014.

\end{thebibliography}
}

\end{document}